\RequirePackage{amsmath}
\documentclass{llncs}
\usepackage{graphicx}
\usepackage{color}
\usepackage{soul}
\usepackage{hyperref}
\usepackage{url,footnote}
\usepackage{dirtytalk}
\usepackage[utf8]{inputenc}
\usepackage[english]{babel}
\usepackage{sidecap}
\usepackage{wrapfig}
\usepackage{multirow}
\usepackage[table,xcdraw]{xcolor}
\usepackage{listings}
\usepackage{amsmath}
\usepackage{amssymb}
\usepackage[linesnumbered,ruled,lined,boxed,commentsnumbered]{algorithm2e}
\usepackage{ifsym}
\usepackage{stmaryrd}
\usepackage{booktabs}
\usepackage[bottom]{footmisc}

\usepackage{booktabs} % For formal tables
\usepackage{silence}
\usepackage{tabularx}
\usepackage{comment}
\usepackage{etoolbox}
\usepackage{hyperref}
\usepackage{listings}
\usepackage{fancybox}
\usepackage{wrapfig}
\usepackage{graphicx}
\usepackage{url}
\usepackage{paralist}
\usepackage{multirow}
\usepackage[flushleft]{threeparttable}
\usepackage{fancybox}
\usepackage{listings}
\usepackage{xspace}

\usepackage{graphicx}
\usepackage{url}
\usepackage{paralist}
\usepackage{multirow}
\usepackage[flushleft]{threeparttable}
\usepackage{fancybox}
\usepackage{listings}
\usepackage{xspace}
\newcommand*\samethanks[1][\value{footnote}]{\footnotemark[#1]}

\usepackage{float}
\usepackage{pgf}
\usepackage{ifthen} 
\usepackage{etoolbox}
\patchcmd{\thebibliography}{\chapter*}{\textbf}{}{}

\newcolumntype{L}[1]{>{\raggedright\let\newline\\\arraybackslash\hspace{0pt}}m{#1}}

\definecolor{mygreen}{HTML}{00CC00}

\newcommand{\myurl}[1]{{\fontsize{8}{8}\selectfont{\url{#1}}}}

\newcommand{\myequationfont}{
\fontsize{8}{9}\selectfont
}

\newcommand{\mycodefont}{
\fontsize{8}{9}\selectfont\ttfamily
}

\usepackage{float}
\usepackage{xparse}   % http://ctan.org/pkg/xparse
\NewDocumentCommand{\rot}{O{90} O{1em} m}{\makebox[#2][l]{\rotatebox{#1}{#3}}}%
\graphicspath{{}}

\definecolor{codegreen}{rgb}{0,0.6,0}
\definecolor{codegray}{rgb}{0.5,0.5,0.5}
\definecolor{codepurple}{rgb}{0.58,0,0.82}
\definecolor{backcolour}{rgb}{0.95,0.95,0.92}
 
\lstdefinestyle{mystyle}{
    backgroundcolor=\color{backcolour},   
    commentstyle=\color{codegreen},
    keywordstyle=\color{magenta},
    numberstyle=\tiny\color{codegray},
    stringstyle=\color{codepurple},
    basicstyle=\footnotesize,
    breakatwhitespace=false,         
    breaklines=true,                 
    captionpos=b,                    
    keepspaces=true,                 
    numbers=left,                    
    numbersep=5pt,                  
    showspaces=false,                
    showstringspaces=false,
    showtabs=false,                  
    tabsize=2
}
 
\hypersetup{
    colorlinks=false,
    linkcolor=blue,
    filecolor=blue,      
    urlcolor=cyan,
}

\usepackage[textsize=tiny]{todonotes} % use this to see SMALL : ) comments
\begin{document}

%\title{Towards supporting Query interoperability between RDF and Graph databases via GREMLINATOR}
% \title{Impact of Context-aware end to end Entity Linking on Wikidata Knowledge Graph}
\title{Encoding Knowledge Graph Entity Aliases in Attentive Neural Network for Wikidata Entity Linking
}
%\title{Linked Data-based Question Answering Benchmark Generation Framework}

%\titlerunning{QaldGen: Personalised Benchmark Generator for Question Answering over Linked Data}
 \author{
 Isaiah Onando Mulang' \inst{1}  \and 
  Kuldeep Singh\inst{2}\samethanks \and
  Akhilesh Vyas\inst{3}\samethanks \and 
 Saeedeh Shekarpour \inst{4}  \and
    Maria-Esther Vidal\inst{3} \and 
     Jens Lehmann\inst{1} \and 
  Soren Auer\inst{3}
 }
\authorrunning{Mulang' Onando et al.}

 \institute{
    Fraunhofer IAIS, University of Bonn, Germany \email{isaiah.mulang.onando,jens.lehmann@iais.fraunhofer.de} \and
  Cerence GmbH, and Zerotha Research, Germany, \email{kuldeep.singh1@cerence.com} \and 
    TIB, Hannover, Germany \email{maria.vidal,akhilesh.vyas@tib.eu} \and 
  University of Dayton, USA, \email{saeedeh@knoesis.org}
 }
\maketitle

%\paragraph{%
%\textbf{Resource Type:} Evaluation benchmarks or Methods  \\
%\textbf{Repository:} \url{https://github.com/dice-group/qald-generator} \\
%\textbf{License:} GNU General Public License v3.0
%}

\begin{abstract}
%The Entity Linking (EL) approaches have been a long-standing research field and find applicability in various use cases such as semantic search, text annotation, question answering, etc. Although effective and robust, current approaches are still limited to particular knowledge repositories (e.g. Wikipedia) or specific knowledge graphs (e.g. Freebase, DBpedia, and YAGO).  
The collaborative knowledge graphs such as Wikidata excessively rely on the crowd to author the information. Since the crowd is not bound to a standard protocol for assigning entity titles, the knowledge graph is populated by non-standard, noisy, long or even sometimes awkward titles. The issue of long, implicit, and nonstandard entity representations is a challenge in Entity Linking (EL) approaches for gaining high precision and recall. Underlying KG in general is the source of target entities for EL approaches, however, it often contains other relevant information, such as aliases of entities (e.g., Obama and Barack Hussein Obama are aliases for the entity Barack Obama).
EL models usually ignore such readily available entity attributes.
In this paper, we examine the role of knowledge graph context on an attentive neural network approach for entity linking on Wikidata. Our approach contributes by exploiting the sufficient context from a KG as a source of background knowledge, which is then fed into the neural network. This approach demonstrates merit to address challenges associated with entity titles (multi-word, long, implicit, case-sensitive).
Our experimental study shows $\approx$8\% improvements over the baseline approach, and significantly outperform an end to end approach for Wikidata entity linking. 
%This work opens a new direction for the research community to pay attention to developing context-aware EL approaches for  collaborative knowledge graphs. 
\end{abstract} 

%%%%%%%%%%%%%%%%%%%%%%%%%%%%%%%%%%%%%%%%%%%%%%%%%%%%%%%%%%%%%%%%%%%%%%%%%%%%%%%%%%%%%
%%%%%%%%%%%%%%%%%%%%%%%%%%%%%%%%%%%%%%%%%%%%%%%%%%%%%%%%%%%%%%%%%%%%%%%%%%%%%%%%%%%%%
%%%%%%%%%%%%%%%%%%%%%%%%%%%%%%%%%%%%%%%%%%%%%%%%%%%%%%%%%%%%%%%%%%%%%%%%%%%%%%%%%%%%%
%%%%%%%%%%%%%%%%%%%%%%%%%%%%%%%%%%%%%%%%%%%%%%%%%%%%%%%%%%%%%%%%%%%%%%%%%%%%%%%%%%%%% 
        %%%%%%%%%%%%%%%%%%%%%%%%%%%%%%%%%%%%%%%%%%%%%%%%%%%%%%%%%%%%%%%%%%%%%%%%%%%%%%%%%%%%%
\section{Introduction}
Entity linking (EL) over Web of data often referred as Named Entity Disambiguation (NED) or Entity Disambiguation is a long-standing field of research in various research communities such as information retrieval, natural language processing, semantic web, and databases since early approaches in 2003 \cite{Balog2018}. 
EL generally comprises two subtasks: \emph{entity recognition} that is concerned with the identification of entity surface forms in the text, and \emph{entity disambiguation} that aims at linking the surface forms with structures and semi-structured knowledge bases (e.g. Wikipedia), or structured knowledge graphs (e.g. DBpedia~\cite{DBLP:conf/semweb/AuerBKLCI07}, Freebase \cite{DBLP:conf/aaai/BollackerCT07} or Wikidata \cite{DBLP:conf/www/Vrandecic12}). \\
\textbf{Research Objectives, Approach, and Contribution.} 
Uniqueness of Wikidata is that the contents are collaboratively edited. As at April 2020; Wikidata contains 83,151,903 items and a total of over 1.2B edits since the project launch\footnote{\url{https://www.wikidata.org/wiki/Wikidata:Statistics}}. User-created entities add additional noise and and non-standard formats since users do not follow a strict naming convention nor a standardised approach ; for instance, there are 1788134 labels in which each label matches with at least two different URIs.
The previous approaches for EL \cite{DBLP:conf/aaai/RaimanR18,sakor2019old} on the textual content consider the well-established knowledge bases such as Wikipedia, Freebase, YAGO \cite{DBLP:conf/www/SuchanekKW07}, and particularly DBpedia. 
%EL approaches over other KGs along with few recently released EL approaches over Wikidata are commonly evaluated over standard datasets (e.g. WikiDisamb30 \cite{DBLP:conf/cikm/FerraginaS10}). These datasets do not contain Wikidata KG specific challenges such as very long entity labels, noisy implicit entities, etc. (cf. section \ref{sec:motivation}).   Furthermore, nearly all (except T-Rex \cite{DBLP:conf/lrec/ElSaharVRGHLS18}) of existing datasets are tailored to Wikipedia (or DBpedia) (e.g., WikiDisamb30 \cite{DBLP:conf/cikm/FerraginaS10}) and may also result into less attention towards Wikidata specific challenges.
%Comparing to other KGs, Wikidata is more challenging for the task of EL since it often contains long entity labels, noisy implicit entities, etc. (cf. section \ref{sec:motivation}). 
%The existing approaches do not embrace the challenges of Wikidata and confine themselves to simplified scenarios or employ a complementary KGs (DBpedia, Yago) \cite{Balog2018} which is a significant subset of Wikidata.
Thereby, Wikidata as the core background KG along with its inherent challenges, has not been studied particularly for the task of EL. %This deficiency might come from its recent emergence, noisy nature, or lack of a training dataset for a long while.

Besides the vandalism and noise in underlying data of Wikidata, collaborative editing of its content adds several aliases of the entities and its description as entity properties (attributes). This enables Wikidata as a rich source of additional information which may be useful for EL challenges. Thus, in this work, we analyse the impact of additional context from Wikidata on Attentive Neural Networks (NN) for solving its entity linking challenges. We develop a novel approach called Arjun, first of its kind to recognise entities from the textual content and link them to equivalences from Wikidata KG. An important strength of Arjun is \underline{an ability to link non-Wikipedia entities of Wikidata} by exploiting unique characteristic of the Wikidata itself (i.e. availability of entity aliases as explained in Section \ref{sec:motivation}). Please note, focus of this paper is not to propose a black-box deep learning approach for entity linking using latest deep learning models such as transformers or graph neural networks. 
In this paper we hypothesise that even though Wikidata is noisy and challenging, but its special property to provide aliases of entities can help an NN better understand the context of the potential entities.
%Our focus in this paper is to understand the impact of KG context on a NN for addressing challenges of the Wikidata KG entity linking.Wikidata is noisy and challenging, but its special property to provide aliases of entities can help an NN better understand the context of the potential entities.  
Since the concept of informing a neural network using contextual data from a KG is our proposed-solution in this work, we believe that traditional neural networks make it more transparent to understand the impact of KG context. Hence, our approach contributes to model attentive neural networks respecting the contextual content and trained on a sizable dataset. In particular, Arjun is a pipeline of two attentive neural networks coupled as follows: 
\begin{enumerate}
    \item In the first step, Arjun utilises a deep attentive neural network to identify the surface forms of entities within the text. 
    \item In the second step, Arjun uses a local KG to expand each surface form from previous step to a list of potential Wikidata entity candidates. Unlike \cite{kolitsas2018end}, Arjun does not use a pre-computed entity candidate list and search entity candidates among all the Wikidata entities.
    \item  Finally, the surface forms, coupled with potential Wikidata candidates, are fed into the second attentive neural network to disambiguate the Wikidata entities further.
\end{enumerate}
Although simple, our approach is empirically powerful and shows $\approx$8\% improvement over the baseline. We also release the source code and all utilised data for reproducibility and reusability on Github\footnote{\url{https://github.com/mulangonando/Arjun}}.
The remainder of the article is structured as follows: section \ref{sec:motivation} motivates our work by discussing Wikidata specific entity linking challenges. Section \ref{sec:relatedwork} discusses related work. This is followed by the formulation of the problem in section \ref{sec:problem}. Section \ref{sec:approach} describes the approach. In section \ref{sec:experiment} we discuss the experimental setup and the results of the evaluation. We conclude in section \ref{sec:conclusion}.

\section{Motivating Examples} \label{sec:motivation}
\begin{figure}
	\centering
	\includegraphics[width=\textwidth]{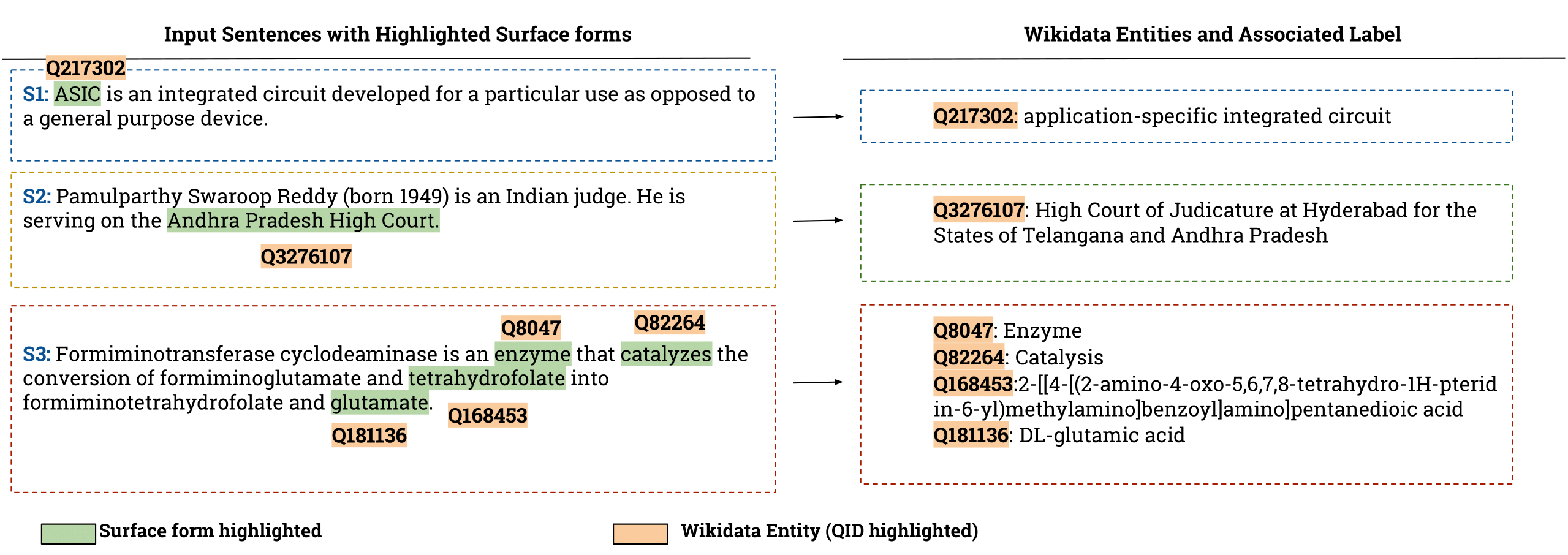}
	\vspace{-4mm}
	\caption{
Wikidata Entity linking Challenges: Besides the challenge of capitalisation of surface forms and implicit nature of entities, Wikidata has several specific challenges, such as very long entity labels and user created entities.  
	}
	\label{fig:motivating}
	\vspace{-4mm}
\end{figure}

We motivate our work by highlighting some challenges associated with linking entities in the text to Wikidata. Wikidata is a community effort to collect and provide an open structured encyclopedic data. The total number of entities described in Wikidata is over 54.1 million~\cite{DBLP:conf/www/Vrandecic12}. Wikidata entities are represented by unique IDs known as \texttt{QID} and \texttt{QIDs} are associated with entity labels. Figure \ref{fig:motivating} shows three sentences extracted from the dataset released by ElSahar et al. \cite{DBLP:conf/lrec/ElSaharVRGHLS18} which aligns 6.2 million Wikipedia sentences to associated Wikidata triples ($<$subject,predicate,object$>$).

In the first sentence S1, the surface form \texttt{ASIC} is linked to a Wikidata entity \texttt{wiki:Q217302} and the entity is implicit (i.e. no exact string match between surface form and entity label). However, ASIC is also known as `Application Specific Integrated Circuit' or Custom Chip. 
Therefore to disambiguate this entity, background information about the surface form will be useful. Please note, we will use this sentence as a running example "Sentence S1". 
In the second sentence S2 the surface form \texttt{Andhra Pradesh High Court} is linked to \texttt{wiki:Q3276107} which has 14 words in the full entity label\footnote{High Court of Judicature at Hyderabad for the States of Telangana and Andhra Pradesh}. 
It is also important to note here that the surface form \texttt{Andhra Pradesh High Court} also contains two sub-surface forms \texttt{Andhra Pradesh} and \texttt{High Court} which are the entity labels of the two Wikidata entities \texttt{wiki:Q1159} and \texttt{wiki:Q671721}. 
An ideal entity linking tool first has to identify \texttt{Andhra Pradesh High Court} as a single surface form then disambiguate the surface form to a long entity label. 
In Wikidata, entity labels and associated aliases can be long (e.g. \texttt{wiki:Q1156234}, \texttt{wiki:Q15885502}).
In addition there are long erroneous entity labels and aliases, such as entity \texttt{wiki:Q44169790}\footnote{\url{https://www.wikidata.org/wiki/Q44169790}} with 62 words in the label and entity \texttt{wiki:Q12766033} with 129 words in one alias. 
Presence of long multi-word entity labels is also specific to Wikidata and poses another challenge for entity linking.  
Furthermore, in sentence S3 illustrated in the Figure \ref{fig:motivating}, the surface form \texttt{tetrahydrofolate} is linked to \texttt{wiki:Q168453} which not only has a multi-word entity label and lowercase surface forms but also contains several numeric and special, non-alphanumeric ASCII characters. Such entities are not present in other public KGs. This is because unlike Wikidata other KGs do not allow users to create new entities and the entity extraction process depends on unique IRIs of Wikipedia pages, WordNet taxonomy, and GeoNames. 
A large number of user-created entities poses specific challenges for entity linking. 
Therefore, it is evident that in addition to generic entity linking challenges such as the impact of capitalisation of surface forms and the implicit nature of entities which are tackled up to a certain extent by approaches for entity linking over Wikipedia and DBpedia \cite{sakor2019old}, Wikidata adds some specific challenges to the entity linking problem.

\section{Related Work} \label{sec:relatedwork}
Several comprehensive surveys exist that detail the techniques employed in entity linking (EL) research; see, for example, \cite{Balog2018}. 
%Likewise, Ji Chen \cite{hengji2019} provides an elaborate reading list for Entity Linking. 
%The review by Shen et al. \cite{shen2015} observes that several EL systems employ a two-step process: Named Entity Recognition (NER) or surface form extraction followed by a linking/disambiguation. 
An elaborate discussion on NER has been provided by Yadav Bethard \cite{Yadav2018ASO}. However, the use of Knowledge Graph as background knowledge for EL task is a relatively recent approach. Here, a knowledge graph is not only used for the reference entities but also offers additional signals to enrich both the recognition and the disambiguation processes. For entity linking, FALCON \cite{sakor2019old} introduces the concept of using knowledge graph context for improving entity linking performance over DBpedia. Falcon creates a local KG fusing information from DBpedia and Wikidata to support entity and predicate linking of questions. We reused the Falcon Background knowledge base and then expand it with all the entities present in the Wikidata (specially non standard entities). 
%Another work in the similar direction is by Seyler et al. \cite{DBLP:conf/acl/SeylerDCHW18}. Authors utilise an extensive set of features as background knowledge to train a linear chain CRF classifier for the NER task. 
%t an approach that leverages background knowledge for improving an underlying deep learning model for EL task. Furthermore, the developments in Deep Learning has introduced a range of models that carry out both NER and NED as a single end to end step \cite{DBLP:conf/emnlp/GaneaH17}.

The developments in deep learning has introduced a range of models that carry out both NER and NED as a single end to end step using various neural network based models \cite{kolitsas2018end}. Kolitsas et al. \cite{kolitsas2018end} enforces during testing that gold entity is present in the potential list of candidates, however, Arjun doesn't have such assumption and generates entity candidates on the fly. This is one reason Arjun is not compared with Kolitsas's work in the evaluation section. Please note, irrespective of the model opted for entity linking, the existing EL approaches and their implementations are commonly evaluated over standard datasets (e.g. CoNLL (YAGO) \cite{DBLP:conf/emnlp/HoffartYBFPSTTW11}). These datasets contain standard formats of the entities commonly derive from Wikipedia URI label. Recently, researchers have explicitly targeted EL over Wikidata by proposing new neural network-based approach \cite{cetoli2019neural}. Contrary to our work, authors assume entities are recognised (i.e. step 1 of Arjun is already done), inputs to their model is a “sentence, one wrong Wikidata Qid, one correct Qid” and using an attention-based model they predict correct Qid in the sentence- more of a classification problem. Hence, 91.6 F-score in Cetoli et al.'s work \cite{cetoli2019neural} is for linking correct QID to Wikidata, given the particular inputs. Their model is not adaptable for an end to end EL due to input restriction.  OpenTapioca \cite{delpeuch2019opentapioca}) is an end to end EL approach to Wikidata that relies on topic similarities and local entity context, but ignores the Wikidata specific challenges (section \ref{sec:motivation}). Works in \cite{sakor2019falcon,mulang2020} are other attempts for Wikidata entity linking.

\section{Problem Statement}\label{sec:problem}
%\todo[inline]{KS to SS: Please modify formalisation, we can put it as variant of entity linking problem as we only target the surface form to entity label linking, ignoring Q value issue. Also, would it not be good to also formalise background knowledge?}
Wikidata is an RDF\footnote{\url{https://www.w3.org/RDF/}} knowledge graph that contains a set of triples $(s, p, o) \in \mathcal{R} \times \mathcal{P} \times (\mathcal{R} \cup \mathcal{L})$, where $\mathcal{R} = \mathcal{C} \cup \mathcal{E} \cup \mathcal{P}$ is the union of all RDF resources. ($\mathcal{C}, \mathcal{P}, \mathcal{E}$ are respectively a set of classes, properties, and entities), and $L$ is the set of literals ($L \cap R = \emptyset$).
An RDF knowledge graph represents a directed graph structure which is formally defined as:

\begin{definition}[Knowledge Graph]
A knowledge graph $KG$ is a directed labelled graph $G(V,E)$, where $V =\mathcal{E} \uplus \mathcal{C} \uplus L$
is a disjoint union of entities $\mathcal{E}$, classes $\mathcal{C}$, and literal values $L$.
The set of directed edges is denoted by $E = \mathcal{P}$, where $\mathcal{P}$ are properties connecting vertices. Please note that there is no outgoing edge from literal vertices.
\end{definition}

\noindent In this paper, we target \textbf{end to end EL task}. The EL for us is defined as recognising the surface forms of entities in the text and then map them to the entities in the background KG. The EL task can be defined as follows:

%\todo[inline]{SS to KS: I added in the following}
\begin{definition}[Entity Linking]
Assume a given text is represented as a sequence of words $w=(w_1,w_2,...,w_N)$ and the set of entities of a KG is represented by set $\mathcal{E}$.
The EL task maps the text into a subset of entities denoted as $ \Theta: w \rightarrow  \mathcal{E'}$ where $\mathcal{E'} \subset \mathcal{E}$.
Herein, the notion of Wididata entity refers to the representation of an entity based on the corresponding label because Wikidata might consider a variety of identifiers (called Q values) for the same label.  
 \end{definition}

The EL task can be divided into two individual sub-tasks. 
The first sub-task Surface Form Extraction is recognising the surface forms of the entities in the text.
This task is similar to Named Entities Recognition (NER). 
However, it disregards identifying the type of entities (e.g. person, place, date, etc.). 

\begin{definition}[Surface form Extraction]
Let $w=(w_1,w_2,...,w_N)$ be a text represented as a sequence of words. 
The surface form extraction is then a function $\theta_1: w \rightarrow  \mathcal{S}$, where the set of surface forms is denoted by $\mathcal{S}=(s_1,s_2,...,s_K)$ ($K\leq N$) and each surface form $s_{x}$ is a sequence of words from start position $i$ to end position $j$: $s_x^{(i,j)}=(w_i,w_{i+1},...,w_j)$. 
\end{definition}

The second sub-task Entity Disambiguation (ED) is mapping each surface form into a set of the most probable entities from the background KG. 

\begin{definition}[Entity Disambiguation]
Let $\mathcal{S}$ be the set of surface forms and $ \mathcal{E}$ the set of entities of the background KG.
Entity Disambiguation is a function $\theta_2: \mathcal{S} \rightarrow  \mathbb{P}(\mathcal{E})$, which assigns a set of entities to each surface form.
\end{definition}
Please note that a single surface form potentially might be mapped into multiple potentially suitable entities.

%\todo[inline]{Saeedeh:are the label of Wikidata in various languages? how did u slice English part? abit unclear}
\section{Arjun: A context-aware Entity Linking Approach} \label{sec:approach}
\begin{figure}
	\centering
	\vspace{-4mm}
	\includegraphics[width=\textwidth]{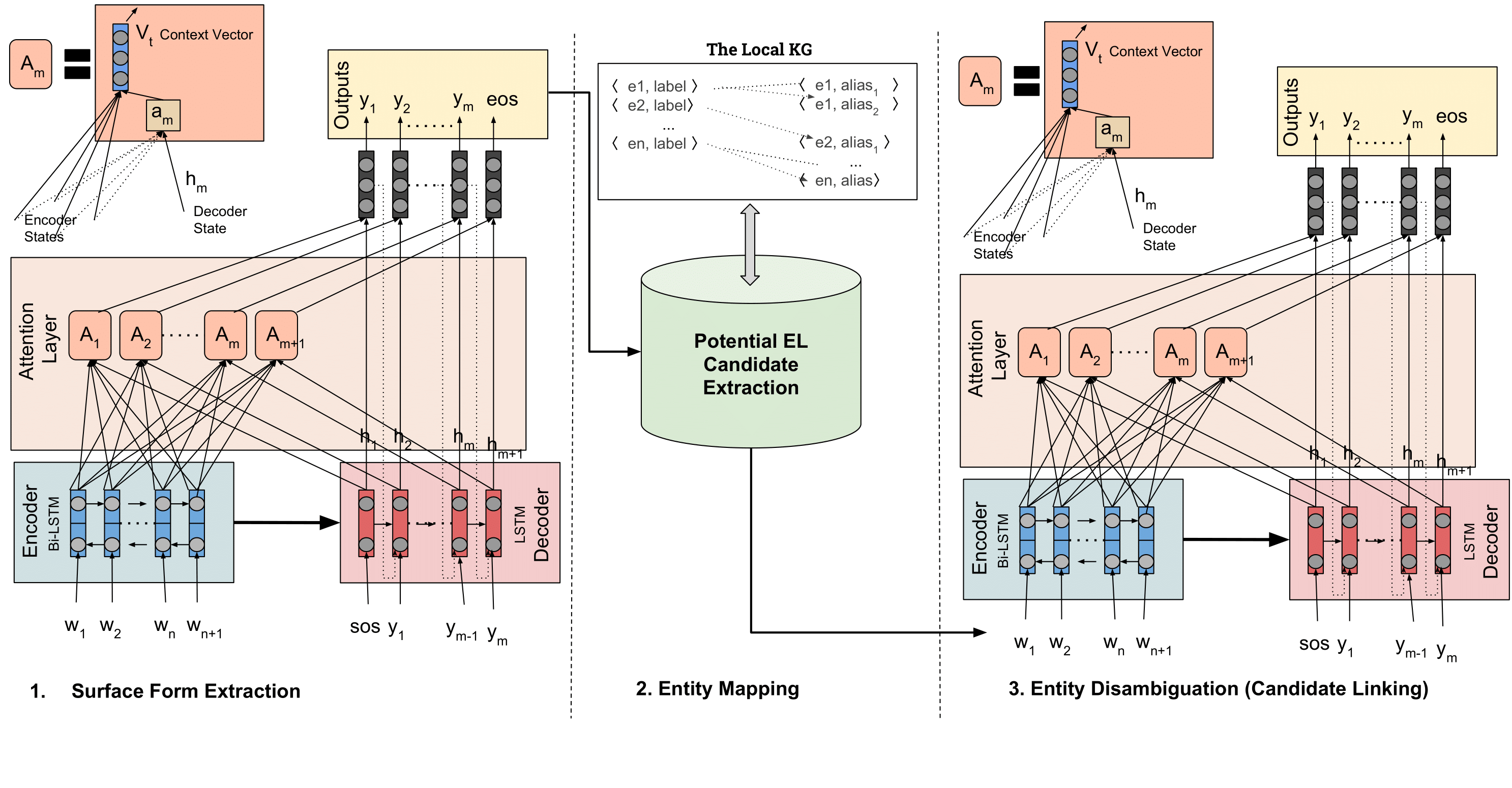}
	\vspace{-4mm}
	\caption{
Proposed Approach Arjun: Arjun consists of three tasks. First task identifies the surface forms using an attentive neural network. Second task induces background knowledge from the Local KG and associate each surface form with potential entity candidates. Third task links the potential entity candidates to the correct entity labels. 
	}
	\label{fig:arch}
	\vspace{-4mm}
\end{figure}
Arjun is illustrated in Figure \ref{fig:arch}. Arjun performs three sub-tasks: 
\begin{enumerate}
    \item surface form extraction which identifies the surface forms of the entities, 
    \item entity mapping (or candidate generation) which maps the surface forms to a list of candidate entities from the Local KG, 
    \item entity disambiguation which selects the most appropriate candidate entity for each surface form.
\end{enumerate}
We devise a context-aware approach based on attentive neural networks for tasks (1) and (3).
We initially introduce our derived Local KG. Then we present the details of our approach for the tasks (1), (2) and (3). 
%We explain our approach below. \\ 
\paragraph{Local KG and Refinement Strategies.} 
Arjun relies on Wikidata as the background knowledge graph. 
Wikidata consists of over 100 million triples in RDF format. 
Wikidata provides dumps of all the entities and associated aliases\footnote{\url{https://dumps.wikimedia.org/wikidatawiki/entities/}}. Although Wikidata has specific challenges for EL, its unique characteristic to provide entity aliases can be utilised in developing an approach for entity linking.
%Each entity also has explicit mention of the language (English, Spanish, Hindi, German, etc.) in the dump. 
Since the training dataset is in English, we extracted all 38.6 million Wikidata entities with English labels and 4.8 million associated aliases from the dumps. We use entity labels and aliases as indexed documents in the Local KG and large portion of it is reused from Local KG built by Sakor et al. \cite{sakor2019old}. 
For example, the entity described in exemplary "Sentence S1" (cf. Figure \ref{fig:motivating}), entity \texttt{wiki:Q217302} with label \textit{application-specific integrated circuit} is enriched in the Local KG with its aliases: ASIC, Custom Chip, and Custom-Chip.

\paragraph{\bf{Model Architecture}}
%\todo[inline]{Saeedeh:there is no support for this statement"However, we propose an extension of entity labels with its aliases (KnownAS) to build a Local KG which has not yet used in literature." How did you add alias name, if there is a publication please add}
For task (1) and (3), our attentive neural model is inspired by the work of Luong et al. \cite{DBLP:conf/emnlp/LuongPM15} and consists of an encoder, a decoder, and an attention layer. We don't claim that an extension of Luong's NN architecture used in this work as novelty, indeed we experiment with already established concepts of LSTM and attentive Neural Networks. We view our attempt of combining these NNs with \textit{background contextual knowledge from a KG} as an interesting perspective for researchers within the community to solve Wikidata KG challenges and is our main novelty. The model in the task (1) is used to identify the surface forms of the entities in the input text. The similar attentive neural model used in the task (3) that selects the most appropriate candidate entity for each surface form (cf. Figure \ref{fig:arch}).

We extended Luong's model by using a Bidirectional Long Short-Term Memory (Bi-LSTM) model for the encoder and one-directional LSTM model for the decoder. 
The input of the encoder is the source text sequence $w=(w_1,w_2,....,w_n,..,w_N)$ where
$w_n$ is the n-th word at time step n and $N$ is the length of the text. 
The encoder encodes the complete sequence and the decoder unfolds this sequence into a target sequence $y=(y_1,y_2,..,y_m,...,y_M)$ where $y_m$ is the m-th word at time step m and $M$ is the length of the target sequence. In our assumption each target sequence ends with EOS (end of sequence) token.
The N and M values can be considered as the last time steps of the source sequence $w$ and the target sequence $y$ respectively. 

Each word of the source sequence is projected to its vector representation acquired from an embedding model $\mathcal{R}^d$ with dimensionality $d$.  
The transformation of the input is represented in the matrix $X$ as:
\begin{equation}
 X=[x_1,x_2,.,x_n,...,x_N]   
\end{equation}
where $x_n$ is a vector with the size $d$ and represents the low dimensional embedding of the word $w_n$.
%%%%%%%%%%%%%%%%%%%%%%%%%%%%%%%%%%%%%%%%%%%%%%%%%%%%%%%%%%%%%%%%%%%%%%%%%%%%%%%%%%%%%%%%%%%%%%%%%%%%%%%%%%
\paragraph{\bf The LSTM Layer:} In our model, the encoder and the decoder consist of single layer of Bi-LSTM and LSTM respectively. Now we explain the LSTM layer.

We model the first layer of our network using a LSTM layer since it has been successfully applied to various NLP tasks. 
Each LSTM unit contains three gates (i.e., input $i$, forget $f$ and output $o$), a hidden state $h$ and a cell memory vector $c$.
The forget gate is a sigmoid layer applied on the previous state $h_{t-1}$ at time step t-1 and the input $x_t$ at time step t to remember or forget its previous state (eq. \ref{eq:forget}).
\begin{equation}
\label{eq:forget}
 f_t=\sigma(W^f[x_t,h_{t-1}]+b^f)   
\end{equation}
Please note that $W$ is the weight matrix and $b$ is the bias vector. 
The next step determines the update on the cell state. 
The input gate which is a sigmoid layer updates the internal value (eq. \ref{eq:input}), and the output gates alter the cell states (eq. \ref{eq:output}).
\begin{equation}
\label{eq:input}
 i_t=\sigma(W^i[x_t,h_{t-1}]+b^i)   
\end{equation}
\begin{equation}
\label{eq:output}
 o_t=\sigma(W^o[x_t,h_{t-1}]+b^o)   
\end{equation}
The next tanh layer computes the vector of a new candidate for the cell state  $\Tilde{C_t}$ (eq. \ref{eq:candidate}). 
Then the old state $C_{t-1}$ is updated by the new cell state $C_t$ via multiplying the old state with the forget gate and adding the candidate state to the input gate (eq. \ref{eq:newcell}). 
The final output is a filtering on the input parts (eq. \ref{eq:output}) and the cell state (eq. \ref{eq:hiddenstate}).
\begin{equation}
\label{eq:candidate}
 \Tilde{C_t}=\tanh(W^C[x_t,h_{t-1}]+b^C)   
\end{equation}
\begin{equation}
\label{eq:newcell}
 {C_t}=f_t\odot C_{t-1}+i_t \odot \Tilde{C_t}   
\end{equation}
\begin{equation}
\label{eq:hiddenstate}
 h_t=o_t\odot \tanh(C_t)  
\end{equation}

where the model learning parameters are weight matrices\\ $W^f,W^i,W^o,W^C$ and bias vectors $b^f,b^i,b^o,b^C$.
The $\sigma$ denotes the element-wise application of the sigmoid function and $\bigodot$ denotes the element-wise multiplication of two vectors.

The Bi-LSTM of the encoder consists of two LSTM layers. 
The first layer takes an input sequence in forward direction (1 to N) and the second layer takes the input in backward direction (N to 1). 
We employ same equations \ref{eq:forget}, \ref{eq:input}, \ref{eq:output}, \ref{eq:candidate}, \ref{eq:newcell}, \ref{eq:hiddenstate} for each LSTM Layer. The final encoder hidden state is produced by the sum of hidden states from both LSTM layers ($h_n = \overrightarrow{h_n} + \overleftarrow{h_n}$) at timestep n.
\paragraph{\bf Attention and Decoder layer.} The decoder layer takes SOS token (start of the sequence) vector and the encoder final states ($h_N$ and $C_N$) as the initial inputs to start decoding source text sequence into a target text sequence. Here we differentiate between the encoder hidden state and decoder hidden state by using the notations $h_n$ at time step n and $h_m$ time step m respectively. Below we explain how the decoder generates target text sequence words $y_m$ one by one.   

In the attention layer, we define attention weights as\\$a_m = [a_{m1},a_{m2},....,a_{mN}]$ for a decoder state at time step m which has the size equals to the number of total time steps in the encoder side. The attention weights contain only scalar values which are calculated by comparing all encoder states $h_n$ and decoder state $h_m$. To calculate the attention weight ($a_{mn}$) of an encoder state at time step n wrt. a decoder state at time step m, we use the following equation (\ref{eq:attnweight}) \cite{DBLP:conf/emnlp/LuongPM15}.

\begin{align}
\label{eq:attnweight}
{a_{mn}} &=\frac{\exp({h}_m \cdot {h}_n)}{\sum{_{n^{'}=1}^{N}}\exp({h}_m \cdot {h_{n'}})}
\end{align}

Where (${h}_m \cdot {h}_n)$ denotes the dot product. The equation \ref{eq:contextVector} computes the context vector $V_m$ as weighted average over all the encoder hidden states ($h_n$) that captures the relevant encoder side information to help in predicting the current target sequence word $y_m$ at time step m and can be defined as:
\begin{align}
    \label{eq:contextVector}
   {V_m} = \sum_{n=1}^{N} {a_{mn}} {{h}_n}
\end{align}

We calculate Attention Vector (${\Tilde{h}_m}$) using the concatenation layer on the context vector ${V_m}$ and decoder hidden state ${h_m}$ for combining information from both the vectors. The equation \ref{eq:ht_new} represent it mathematically (where $\tanh$ is an activation function same as describe in  \cite{DBLP:conf/emnlp/LuongPM15}).

\begin{equation}
\label{eq:ht_new}
    {\Tilde{h}_m} = \tanh{({W_v}[{v_m};{h_m}])}
\end{equation}

Finally, we apply softmax layer on the attention vector ${\Tilde{h}_m}$ for predicting a word of a target text sequence from the predefined vocabulary of the complete target text sequences.
\begin{equation}
\label{eq:softmax_new}
    p(y_m|y_{<m},x) = \text{softmax} ({W_s}{\Tilde{h}_m})
\end{equation}

Where $W_s$ is weight matrix of softmax layer and $p$ is probability. Please note that the decoder stops producing words once it encounters EOS (end of sequence) token or $m$ is equal to M.

\subsection{Entity Mapping Process}
%\todo[inline]{KS to SS: can you please improve below paragraph (if needed)}
%\todo[inline]{SS to KS: we have already the formalization, I revised the text, some parts are very unlcear, e.g., Each surface form associated with contextual information in the form of additional Wikidata entity labels is then passed}
The local KG acts as a source of background knowledge. It is an indexed graph (created using the same methodology proposed by Sakor et al. \cite{sakor2019old} and reusing a large portion of the indexed graph built by the authors) where each entity label is extended with its aliases from Wikidata. Once Task 1 identifies surface forms in the input sentence, the entity mapping step (Task 2) takes each surface form and retrieves all the entities for which entity label(s) in the local KG matches with the surface form.
Next, the full list of the entity candidates is then passed into the Step 3 of Arjun as input to predict (disambiguate) the best Wikidata entity labels.

%%%%%%%%%%%%%%%%%%%%%%%%%%%%%%%%%%%%%%%%%%%%%%%%%%%%%%%%%%%%%%%
Let us trace our approach for the sentence S1 of Figure \ref{fig:motivating} to understand the steps better. The sentence S1 \texttt{ASIC is an integrated circuit developed for particular use as opposed to a general-purpose device} is fed to the attentive neural model comprises of an encoder (Bi-LSTM), decoder (LSTM), and an attention layer as an input for the surface form extraction task. Thereby, the term \texttt{ASIC} is recognised as a surface form.
Then, for the entity mapping task, we populate a Local KG to generate candidate entities associated with this surface form. We employ semantic search (reused from Falcon \cite{sakor2019old}) to identify entity candidate labels for \texttt{ASIC} which returns \texttt{Application Specific Integrated Circuit}. The last step of Arjun is entity disambiguation. In this step, the surface form \texttt{ASIC} along with \texttt{Application Specific Integrated Circuit} is fed into the encoder as the input sequence. Here, we utilise identical attentive neural network used for surface form extraction task. This attentive neural network decides the context of \texttt{ASIC} using extra information in the form of associated alias to correctly link to the the Wikidata entity \texttt{application-specific integrated circuit} (Q217302).

%the output of this step is surface form separated by comma like "enzyme,catalyzes,tetrahydroflorate,glutamate" (sentance S3: cf. Figure \ref{fig:motivating}) that help us to know the boundaries of the surface form.

%\paragraph{Inducing Background Knowledge} 
%FYI.
%we use semantic search to extract entity label per surface form

%\paragraph{Candidate Linking} 
%FYI.
%we pass surface form plus associated entity label. So surface forms act as hidden state just like SINA.
%%%%%%%%%%%%%%%%%%%%%%%%%%%%%%%%%%%%%%%%%%%%
%%%%%%%%%%%%%%%%%%%%%%%%%%%%%%%%%%%%%%%%%
\section{Experimental Setup} \label{sec:experiment}
\subsection{Dataset}
We rely on the recently released T-REx \cite{DBLP:conf/lrec/ElSaharVRGHLS18} dataset that contains 4.65 million Wikipedia extracts (documents) with 6.2 million sentences. 
These sentences are annotated by 11 million Wikidata triples.
In total, over 4.6 million surface forms are linked in the text to 938,642 unique entities. 
T-REx is the only available dataset for Wikidata with such a large number of triple alignment. 
We are not aware of any other dataset explicitly released for Wikidata entity linking challenges. 
Please note that the popular entity linking datasets (e.g. CoNLL (YAGO) \cite{DBLP:conf/emnlp/HoffartYBFPSTTW11}) have linked entities either to Wikipedia, YAGO, Freebase or DBpedia. Work in \cite{cetoli2019neural,delpeuch2019opentapioca} attempt to develop approaches for EL over Wikidata and simply align (map using owl:sameAs) existing Wikipedia based dataset to Wikidata. However, our focus in this paper is to solve Wikidata specific challenges and these datasets do no embrace Wikidata specific challenges for entity linking.
We divide the T-REx dataset into an 80:20 ration for training and testing.
%Hence, T-REx provides us with a large resource of linked Wikidata entities compared to other generic entity linking datasets (e.g. ACE2004 \cite{DBLP:conf/acl/RatinovRDA11}, IITB \cite{Kulkarni:2009:CAW:1557019.1557073}, Microposts2014 \cite{DBLP:conf/lrec/RizzoET14}, LC-QuAD \cite{trivedi2017lc} etc.) which have linked entities either to Wikipedia or DBpedia. 
\subsection{Baseline}
%\todo[inline]{KS to SS: I tried adding why we dont use another SOTA or baseline- can you please explain it better? I am not very convinced with my explanation below}
%\todo[inline]{done, please review}
In this work, we pursue the following research question: ``How well does the attentive neural network perform for entity linking task leveraging background knowledge particularly for a challenging KG such as Wikidata?''
To the best of our knowledge, it is a pioneering work for the task of entity linking on the Wikidata knowledge graph where it considers the inherent challenges (noisy nature, long entity labels, implicit entities). 
Therefore, we do not compare our approach to generic entity linking approaches which typically either do not use any background knowledge or employ the well-established knowledge graphs such as DBpedia, YAGO, Freebase. 
Our approach Arjun comprises all three tasks illustrated in Figure \ref{fig:arch}. To elaborate the advantage of inducing additional context post NER step, we built a "baseline" which is an end to end neural model. The "baseline" in our case is the attentive neural network employed in Task 1 without any background knowledge (or can be seen as end to end EL using attentive neural network).
In fact, in the task (1) (cf. Figure \ref{fig:arch}), the baseline directly maps the text to a sequence of Wikidata entities without identifying surface form candidates. 
Hence, the baseline approach is the modified version of Arjun. With a given input sentence, the baseline implicitly identifies the surface forms and directly links them to Wikidata entities. Unlike Arjun, the baseline does not use any KG context for the expansion of the surface forms. We also compare Arjun with recently released SOTA for Wikidata entity linking- OpenTapioca \cite{delpeuch2019opentapioca} which is an end to end EL approach. We are not aware of any other end to end EL tool/approach released for Wikidata.
%The baseline model accepts text sequence as input and directly link the entities to the Wikidata entities. 

\subsection{Training Details}

\paragraph{Implementation details}
We implemented all the models using the PyTorch framework. 
The local KG and the semantic search is implemented using Apache Lucene Core\footnote{\url{https://lucene.apache.org/core/}} and Elastic search \cite{gormley2015elasticsearch}. The semantic search returns entity candidates with a score (higher is better). We reuse the implementation of Falcon local KG \cite{sakor2019old} for the same. After empirically observing the performance, we set the threshold score to 0.85 for selecting the potential entity candidates per surface form (i.e. the parameter is optimised on the test set).
We reused pre-trained word embeddings from Glove \cite{DBLP:conf/emnlp/PenningtonSM14} for the attention based neural network.
These embeddings have been pre-trained on Wikipedia 2014 and Gigaword 5\footnote{\url{https://nlp.stanford.edu/projects/glove/}}. 
We employ 300-dimensional Glove word vectors for the training and testing of Arjun. The models are trained and tested on two Nvidia GeForce GTX1080 Ti GPUs with 11GB size. Due to brevity, detailed description of training details can be found in our public Github.

\paragraph{Dataset Preparation}
%Limited computing resources restrict us to keep the input text sequence length up to 25 words. It is due to the high computational requirements for global attention models. 
We experimented initially with higher text sequence lengths but resorted to 25 words due to GPU memory limitation. 
In total, we processed 983,257 sentences containing 3,133,778 instances of surface forms (not necessarily unique entities) which are linked to 85,628 individual Wikidata entities. 
From these 3,133,778 surface forms occurrences, approximately 62\% \underline{do not} have exact match with a Wikidata entity label. 
%We perform some prepossessing (e.g., tokenisation, normalisation) of the complete T-REx dataset. 
%We also introduce a separation token between the entities of the target sequence and replace the year, number, and out of vocabulary words in all sequences with the unique identifiers such as YEAR, NUMBER, UNKNOWN tokens.

\subsection{Results}

Table \ref{tab:model_result} summarises performance of Arjun compared to the baseline model and another NED approach. 
We observe nearly 8\% improvement in the performance over baseline and Arjun significantly outperforms another end to end EL tool OpentTapioca. Arjun and OpenTapioca generate entity candidates on the fly, i.e., out of Millions of Wikidata entities, the task here is to reach to top-1 entity. This contrasts with other end to end entity linking approaches such as \cite{kolitsas2018end}, which rely on a pre-computed list of 30 entity candidates per surface form. This translates into extra complexity due to the a large search space for generating entity candidates in the case of Arjun.
Our solution demonstrates a clear advantage of using KGs as background knowledge in conjunction with a attention neural network model. 
We now detail some success and failure cases of Arjun.
~\\

\begin{table}[hb!]
	\centering
	\caption{Performance of Arjun compared to the Baseline.}
	\resizebox{0.55\columnwidth}{!}{%
        \begin{tabular}{ l l l l }
    	    \toprule
            \textbf{Method} & \textbf{Precision} & \textbf{Recall} & \textbf{F-Score} \\
            \midrule
            {\it baseline}& 0.664 & 0.662 & 0.663 \\
            {\it OpenTapioca \cite{delpeuch2019opentapioca}}& 0.407 & 0.829 & 0.579 \\
            {\it Arjun}& \underline{0.714} & \underline{0.712} & \underline{0.713} \\
            \bottomrule
        \end{tabular}
        }
    \label{tab:model_result}
\end{table} 

\paragraph{\textbf{Success Cases of Arjun}}
Arjun achieves 0.77 F-Score for the surface form extraction task. 
Arjun identifies the correct surface form for our exemplary sentence S1 (i.e. \texttt{ASIC}) and links it to the entity label \textit{Application Specific Integrated Circuit} of \texttt{wiki:Q217302}. 
The baseline can not achieve the linking for this sentence. 
In the Local KG, the entity label of \texttt{wiki:Q217302} is enriched with aliases that also contain \texttt{ASIC}. 
This allows Arjun to provide the correct linking to the Wikidata entity containing the long label. 
Background knowledge induced in the attentive neural network also allows us to link several long entities correctly. 
For example, in the sentence "The treaty of London or London convention or similar may refer to," the gold standard links the surface form \texttt{London convention} with the label \textit{Convention on the Prevention of Marine Pollution by Dumping of Wastes and Other Matter} (c.f. \texttt{wiki:Q1156234}). The entity label has 14 words, and Arjun provides correct linking. OpenTapioca on the other hand have high recall(it has high number of False Positives), however, the precision is relatively quite low. The limited performance of OpenTapioca was due to the fact that it finds limitation in linking non Wikipedia entities which constitute a major portion in the dataset. This demonstrate strength of Arjun in also linking non standard, noisy entities which are not part of Wikipedia.
\paragraph{\textbf{Failure Cases of Arjun}}
In spite of the successful empirical demonstration of Arjun, we have a few types of failure cases. 
For example in the sentence: `Two vessels have borne the name HMS Heureux, both of them captured from the French' has two gold standard entities (\texttt{Heureux} to \textit{French ship Heureux} (\texttt{wiki:Q3134963}) and \texttt{French} to \textit{French} (\texttt{wiki:Q150})). 
Arjun links \texttt{Heureux} to \textit{L'Heureux} (\texttt{wiki:Q56539239}). This issue is caused by the semantic search over the Local KG while searching for the potential candidates per surface form. 
In this case, L'Heureux is also returned as one of the potential entity candidates for the surface form \texttt{Heureux}.  
A similar problem has been observed in correctly mapping the surface form \texttt{Catalan} to \texttt{wiki:Q7026} (\textit{Catalan Language}) where Arjun links \texttt{Catalan} to Catalan (\texttt{wiki:Q595266}). 
Another form of failure case is when Arjun identifies and links other entities which are not part of the gold standard. 
The sentence `Tom Tailor is a German vertically integrated lifestyle clothing company headquartered in Hamburg' has two gold standard entity mappings: \texttt{vertically integrated} to \textit{vertical integration} (\texttt{wiki:Q1571520} and \texttt{Hamburg} to \textit{Hamburg} (\texttt{wiki:Q1055}). 
Arjun identifies \textit{Tom} (\texttt{wiki:Q3354498}) and \textit{Tailor} (\texttt{wiki:Q37457972}) as the extra entities and can not link \texttt{vertically integrated}.  
For brevity, a detailed analysis of the failure cases per entity type (very long label, noisy nonstandard entity), performance loss due to semantic search can be found in our Github.
\paragraph{\textbf{Limitations and Improvements for Arjun}}
Arjun is the first step towards improving a deep learning model with additional contextual knowledge for EL task. Arjun can be enhanced in various directions considering current limitations. We list some of the immediate future extensions: 
\begin{enumerate}
    \item \emph{Enhancing Neural Network with Multiple layers:} Arjun currently has a Bi-LSTM and a single layer LSTM for the encoder and the decoder respectively. 
    It has been empirically observed in sequence to sequence models for machine translations that the models show significant improvements if stacked with multiple layers \cite{DBLP:conf/aclwat/ShuM16}. 
    Therefore, with more computing resources, the neural network model used in Arjun can be enhanced with multiple layers.
    \item \emph{Alternative Models:} In this article, our focus is to empirically demonstrate how background knowledge can be used to improve an attentive neural network for entity linking. 
    Several recent approaches \cite{DBLP:conf/nips/VaswaniSPUJGKP17,DBLP:conf/emnlp/EdunovOAG18,DBLP:journals/corr/abs-1810-04805} enhance the performance NER and can be used in our models for task (1) and task (3). 
    \item \emph{Improving NER:} there is a room of improvement regarding surface form extraction where Arjun currently achieves an F-score of 0.77. The latest context-aware word embeddings \cite{DBLP:journals/corr/abs-1803-03665} can be re-used in Arjun or completely replacing NER part with latest language models such as BERT \cite{DBLP:journals/corr/abs-1810-04805}. 
    \item \emph{Replacing Semantic Search:} Another possibility of improvement is in the second step of our approach (i.e., inducing background knowledge). Currently, we rely on \underline{very trivial} semantic search (same as \cite{sakor2019old}) over the Local KG to extract Wikidata entity candidates per surface form. Ganea et al. \cite{DBLP:conf/emnlp/GaneaH17} developed a novel method to embed entities and words in a common vector space to provide a context in an attention neural network model for entity linking. This approach could potentially replace semantic search. Classification is seen as one of the most reasonable and preferred ways to prevent out of scope entity labels \cite{kolitsas2018end}. On the contrary, Sakor et al. \cite{sakor2019old} illustrated that expanding the surface forms the way we did, works pretty well for short text. Our hypothesis was that it should also work for Arjun, which is not completely true if we see our empirical results. Hence, in this paper, we do not claim that every step we took was the best, but after our empirical study, we demonstrate that the candidate expansion by Sakor et al. doesn’t work well. However, it solves our purpose of inducing context in the NN which is the main focus of the paper. It leads to an interesting discussion: what is the most efficient way to induce KG context in a NN, maybe the classification one?- one need to empirically prove and we leave it for future work. 
    \item \emph{Coverage restricted to Wikidata:} Effort can be made in the direction to develop common EL approach targeting multiple knowledge graphs with standard and nonstandard entity formats. 
    %\todo[inline]{KS to SS: you had some idea to replace semantic search. Please write it here}
     %This approach could potentially replace semantic search. Applying this approach to find potential candidates in the Local KG could certainly benefit Arjun. 
\end{enumerate}
%\section{Error Analysis}
%%%%%%%%%%%%%%%%%%%%%%%%%%%%%%%%%%%%%%%%%%%%
%%%%%%%%%%%%%%%%%%%%%%%%%%%%%%%%%%%%%%%%%

\section{Conclusion}\label{sec:conclusion}
In this work, we focused on introducing the limitations of EL on Wikidata in general, presented the novel approach Arjun, and outlined deficiencies of Arjun, which in particular will guide future work on this topic. In this work, we empirically illustrate that for a challenging KG like Wikidata, if a model is fused with additional context post-NER step, it improves entity linking performance. However, this work was our first attempt towards a longer research agenda. We plan to extend our contribution particularly in the following directions
(i) extending towards joint entity and predicate linking and use latest language models for NER task, 
(ii) enriching the background KG to several interlinked KG from Linked Open Data (DBpedia, Freebase, YAGO), 
(iii) extending Arjun for the learning entities across languages (currently limited to English).

%%%%%%%%%%%%%%%%%%%%%%%%%%%%%%%%%%%%%%%%%%%%%%%%%%%%%%%%%%%%%%%%%%%%%%%%%%%%%%%%%%%%%
%%%%%%%%%%%%%%%%%%%%%%%%%%%%%%%%%%%%%%%%%%%%%%%%%%%%%%%%%%%%%%%%%%%%%%%%%%%%%%%%%%%%%
%%%%%%%%%%%%%%%%%%%%%%%%%%%%%%%%%%%%%%%%%%%%%%%%%%%%%%%%%%%%%%%%%%%%%%%%%%%%%%%%%%%%%
%%%%%%%%%%%%%%%%%%%%%%%%%%%%%%%%%%%%%%%%%%%%%%%%%%%%%%%%%%%%%%%%%%%%%%%%%%%%%%%%%%%%%  %%%%%%%%%%%%%%%%%%%%%%%%%%%%%%%%%%%%%%%%%%%%%%%%%%%%%%%%%%%%%%%%%%%%%%%%%%%%%%%%%%%%%

%========= ACKNOWLEDGEMENTS============
\section{Acknowledgments} 
  This work is Co-funded by the
European Union’s Horizon 2020 research and innovation programme under the QualiChain Project, Grant Agreement No. 822404; and the IASIS project, Grant Agreement No. 727658.
 
\bibliographystyle{plain}
\bibliography{main}

\end{document}